\def\year{2020}\relax
\pdfoutput=1

\documentclass[letterpaper]{article} 
\usepackage{ifpdf} 
\usepackage{aaai20}  
\usepackage{times}  
\usepackage{helvet} 
\usepackage{courier}  
\usepackage[hyphens]{url}  
\usepackage{graphicx} 
\usepackage{graphicx}  
\usepackage{latexsym}
\usepackage{mathtools}
\usepackage{array} 
\usepackage{booktabs}  
\usepackage{threeparttable}  
\usepackage{multicol}  
\usepackage{multirow}  
\usepackage{amsmath}
\usepackage{amsfonts}
\usepackage{epsfig}
\usepackage{algorithm}
\usepackage{algorithmic}

\title{Semantic Graph Convolutional Network \\ for Implicit Discourse Relation Classification}

\author{Yingxue Zhang\textsuperscript{\rm 1}, \Large \textbf{Ping Jian\textsuperscript{\rm 1}, Fandong Meng\textsuperscript{\rm 2}, Ruiying Geng\textsuperscript{\rm 1}, Wei Cheng\textsuperscript{\rm 1}, Jie Zhou\textsuperscript{\rm 2}}\\ 
\textsuperscript{\rm 1}Beijing Institute of Technology, China\\ 
\textsuperscript{\rm 2}Pattern Recognition Center, WeChat AI, Tencent Inc, China\\
\{zhangyingxue, pjian, rygeng, wcheng\}@bit.edu.cn\\
\{fandongmeng, withtomzhou\}@tencent.com 
}
\begin{document}
\ifpdf

\maketitle

\begin{abstract}
Implicit discourse relation classification is of great importance for discourse parsing, but remains a challenging problem due to the absence of explicit discourse connectives communicating these relations. Modeling the semantic interactions between the two arguments of a relation has proven useful for detecting implicit discourse relations. However, most previous approaches model such semantic interactions from a shallow interactive level, which is inadequate on capturing enough semantic information. In this paper, we propose a novel and effective Semantic Graph Convolutional Network (SGCN) to enhance the modeling of inter-argument semantics on a deeper interaction level for implicit discourse relation classification. We first build an interaction graph over representations of the two arguments, and then automatically extract in-depth semantic interactive information through graph convolution. Experimental results on the English corpus PDTB and the Chinese corpus CDTB both demonstrate the superiority of our model to previous state-of-the-art systems.  
\end{abstract}

\section{Introdution}
\noindent Usually, sentences in a text are not isolated but semantically connected with discourse relations. Therefore, identifying discourse relations could benefit many downstream NLP applications such as question answering~\cite{Liakata2013ADC}, machine translation~\cite{Li2014AssessingTD,Xiong2019DNMT}, text summarization~\cite{Gerani2014AbstractiveSO} and so forth.

As shown in Table 1, there are two branches for the discourse relation classification task: one is the explicit discourse relation classification where relations rely on connectives, e.g. `however' and `because', and the other is the implicit discourse relation classification without the guidance of connectives. Previous studies treat explicit discourse relation recognition as a discourse connective disambiguation problem and have achieved pretty good performance with F1 scores higher than 90\%. However, implicit discourse relation detection can be much challenging since we can only infer the logical relations through the deep semantics hidden in the texts with no connectives to rely on. This makes it the current bottleneck for building an active discourse parser. 

\begin{table} [t]
\centering
\scalebox{1.0} {
\begin{tabular}{ll}
\hline
\textbf{\multirow{2}*{Explicit}} 
& \textit{Arg1: We're standing in gasoline}\\
& \textit{Arg2: \textbf{So} don't smoke}\\ 
\hline
\textbf{\multirow{2}*{Implicit}} 
& \textit{Arg1:Heating oil prices also \textbf{rose}}\\
& \textit{Arg2:November gasoline \textbf{slipped} slightly} 
\\ 
\hline
\end{tabular} 
}
\caption{Real cases from PDTB$2.0$. The word `So' in the explicit case is a connective triggering a `Cause' relation. And interactive semantics provided by the word pair (rose, slipped) in the implicit case will help to trigger a contrast relation.} 
\end{table}

A number of efforts have been done to improve the performance of implicit discourse relation classification. 
Early studies~\cite{Pitler2009AutomaticSP,Lin2009RecognizingID} propose to extract informed linguistic and semantic features from texts and design machine learning algorithms, where word pairs are heavily used as the semantic interactive features. These features do play a role to some extent but are limited by problems of data sparsity~\cite{Biran2013AggregatedWP} and semantic gap~\cite{Zhao2002NarrowingTS}. With the developing of deep learning, neural networks have shown outstanding performance on sentence modeling. Some end-to-end deep sentence modeling based approaches have advanced the performance of discourse relation classification. Some researchers~\cite{ji2015one,Wang2016TwoES,Qin2017AdversarialCN} propose to learn the semantic representation of each argument with neural networks, such as the CNN, the RNN, and the Bi-LSTM for classification.

These methods effectively capture inner semantic connections of each argument via distributed representation learning.
However, they didn't consider semantic interactions between the two arguments. Different from traditional sentence modeling, discourse relation identification is a bi-sequence problem, and the direct contact between the arguments is expected to play an important role.
Considering the implicit case shown in Table 1, the word pair (rose, slipped) might trigger a contrast relation directly, demonstrating that semantic interactions will help in detecting discourse relations.

Recent studies move one step further and exploit ways to model semantic interactions through the gated relevance network~\cite{Chen2016ImplicitDR}, attention mechanism~\cite{Lan2017MultitaskAN} or feature engineering~\cite{lei2018linguistic}. Although these methods have demonstrated clear benefits of modeling semantic interactions, they so far capture semantic interactions on a shallow unstructured level. How to capture semantics from a deeper interactive level to fully take advantages of the underlying latent semantic structure in the arguments remains a significant challenge.

In this work, we propose a novel and more effective Semantic Graph Convolutional Network (SGCN), to model the semantic interactions between the arguments. 
Firstly, we encode each argument into its positional representations via a bidirectional LSTM~\cite{Hochreiter1997LongSM}. 
Secondly, we build a semantic interaction graph to model the latent inter-argument semantic structure. Nodes of the graph are initialized as the argument's positional representations. Edges connect the nodes from different arguments, with weights indicating the strength of semantic association between these two connected nodes. 
Thirdly, deeper interactive features are extracted from the built semantic interaction graph through graph convolution, where each node has automatically incorporated the semantic information of adjacent nodes along weighted edges.  
Finally, all interactive features are aggregated and fed into a MLP classifier after a Concat-Pooling layer. 

We evaluate our approach on the English benchmark of PDTB 2.0 and the Chinese benchmark of CDTB. Experimental results show that the SGCN outperforms previous best-performing neural models. We also compute the interactive score matrix before and after the GCN respectively for visualization, which demonstrates that the SGCN successfully captures useful interactive semantics.
To the best of our knowledge, this is the first work that exploits a graphical structure to model the semantic interaction between arguments. 
Additionally, we provide a more concise and straightforward way to extract deep interactive features for relation classification.

\begin{figure*}[t!]
\centering
\includegraphics[width=1.0\textwidth]{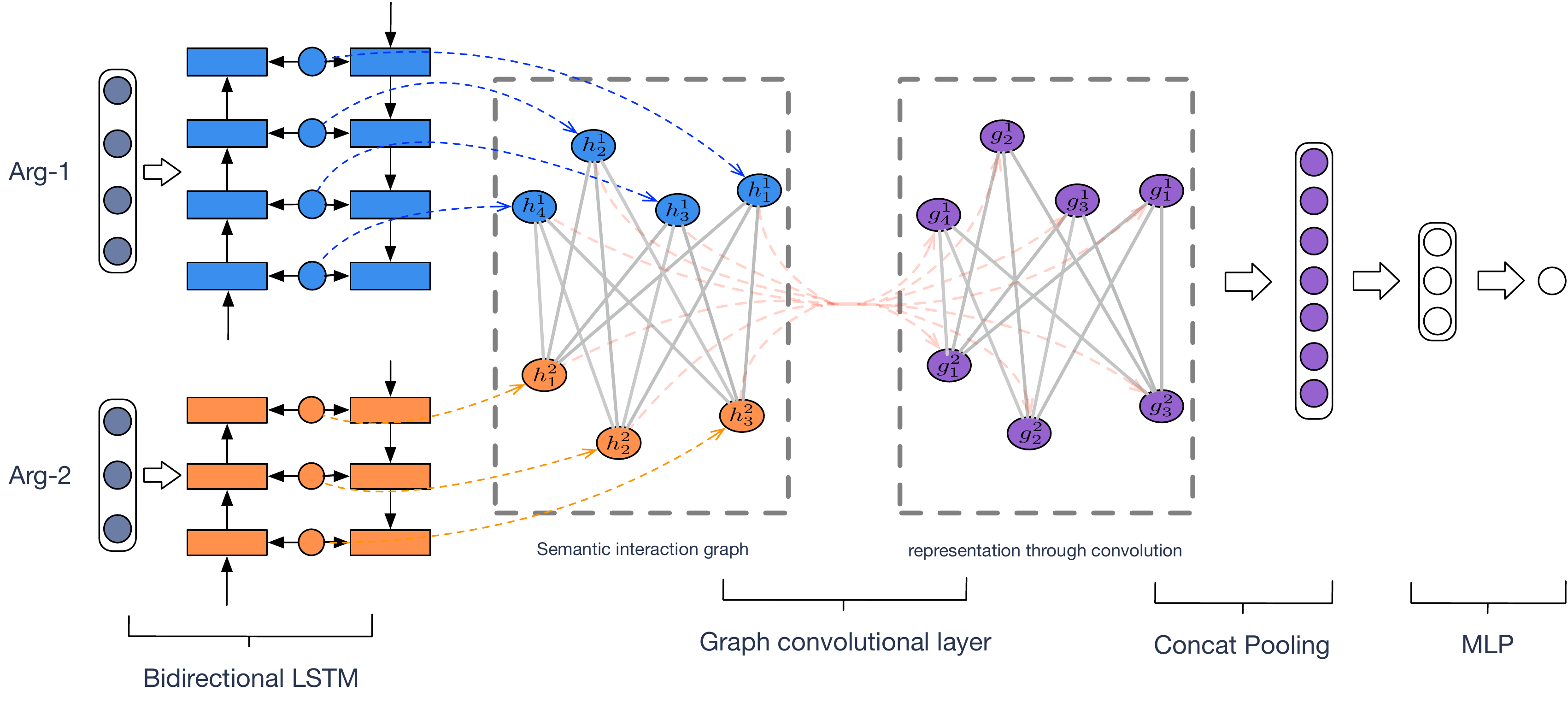}
\caption{The architecture of SGCN. Here, colored dotted lines indicate the correspondence between representations from different layers, and solid gray lines between nodes are edges with weights of the semantic graph.}  
\centering
\label{model architechture}
\end{figure*}

\section{Related Work}
\subsection{Implicit Discourse Relation Classification} 
Both as crucial components for discourse parsing, implicit discourse relation classification is much more challenging than the explicit one due to the absence of connectives. To promote the development of this task, the Penn Discourse Treebank (PDTB) 2.0 was released in 2008~\cite{PrasadPenn}. 
Since then, a surge of studies has been conducted to predict discourse relations, including earlier methods based on surface linguistic features~\cite{Park2012ImprovingID}, and methods aiming to design semantic features like role semantics~\cite{roth2018role}, and end-to-end methods based on sentence modeling~\cite{ji2015one,Rutherford2016RobustNN}. Sentence modeling based methods have earned a much-deserved break in this area but soon reached the bottleneck due to the absence of the semantic interactions between the arguments.

Our work is inspired by recent studies that focus on capturing inter-argument semantics for implicit discourse relation classification.
\citeauthor{Lin2009RecognizingID}~\shortcite{Lin2009RecognizingID}, \citeauthor{She2018LeveragingHD}~\shortcite{She2018LeveragingHD} try to select surface linguistic interactive features between the arguments such as word pairs, 
which are useful to some extent but suffer from the feature sparsity problem~\cite{Biran2013AggregatedWP}. Since surface strings of word pairs are too sparse to work well, researchers consider capturing semantic interactions with neural networks.
~\citeauthor{Rnnqvist2017ARN}~\shortcite{Rnnqvist2017ARN} adopt attention based methods to compute new argument-aware representations as interactive features, but ignore the interactions happening between the semantic elements of each argument, such as word pairs or phrase pairs, which is critical for this task.
~\citeauthor{Chen2016ImplicitDR}~\shortcite{Chen2016ImplicitDR} propose a gated relevance network using a relevance matrix to capture semantic interactions between word pairs. 
~\citeauthor{lei2018linguistic}~\shortcite{lei2018linguistic} encode the semantic interaction, topic continuity, and attribution as combined features and feed them into a Naive Bayes classifier for classification, which requires lots of complex feature engineering work. 

Different from these works, a semantic interaction graph is built in our methods to model the latent semantic interaction structure between the arguments, to be more precise, between the semantic components of each argument. Additionally, interactive features are filtered out automatically through a graph convolutional operation on the structural interaction graph, which is in a deeper level than those previous methods.

Some researchers try to improve the performance of implicit discourse relation classification via utilizing external resources, such as the explicit discourse relation corpus~\cite{dai2018improving}, other corpora on similar tasks~\cite{Liu2016ImplicitDR,Lan2017MultitaskAN} or additional annotations, e.g. implicit connectives~\cite{Qin2017AdversarialCN}. 
It's worth noting that our model does not use any other corpus or external annotation information, while achieves comparable performance. 

\subsection{Graph Convolutional Network}
The Graph Convolutional Network (GCN) was first proposed by \citeauthor{kipf2016semi}~\shortcite{kipf2016semi}, and achieves state-of-the-art classification results on a number of benchmark graph datasets. As a special form of Laplacian smoothing, the GCN model computes new representations of nodes as the weighted average of itself and its neighbors~\cite{Li2018DeeperII}. Further studies explore GCN in many NLP tasks such as machine translation~\cite{bastings2017graph}, semantic role labeling~\cite{Marcheggiani2017EncodingSW} and text classification~\cite{yao2018graph}. In these studies, GCN is used to encode syntactic structure of sentences. 
None of them focuses on incorporating semantic information with the graph.  
In this work, we first employ GCN to encode interactive semantics between two arguments. When building the graph, the argument's positional representations, who carry inner semantics, are regarded as nodes, and semantic relations are calculated as the weights of edges.

\section{Methodology}
In this section, we present the SGCN in detail. The framework of the model is depicted as Figure \ref{model architechture}. It consists of four layers: 
\textbf{(1)} The encoding layer, where we encode each argument into vectors by a Bi-LSTM. And the output of Bi-LSTM at each step is viewed as a positional representation of this argument, who carries inner semantics. \textbf{(2)} The semantic interaction layer, where we build the semantic interaction graph to model the latent semantic structure between the two arguments. \textbf{(3)} The graph convolutional layer, where we extract more in-depth interactive features from the built graph through graph convolution. \textbf{(4)} The classification layer, where we feed the deeper interactive features into a MLP classifier after Concat-Pooling. In the following of this section, we will describe the details of these parts. 
\subsection{Encoding Layer}
Each of the arguments is encoded into vectors by a Bidirectional Long Short-Term Memory network (Bi-LSTM) to incorporate context information.
Here is a brief introduction of LSTM. Given a variable-length sequence S = ($x_1,x_2,...,x_T$), where $x_t$ represent the \textit{t}-th word embedding. At step \textit{t}, LSTM calculate $h_t$ as follows:
\begin{equation}
\begin{array}{cl}
  i_t=   & \sigma(W_ih_{t-1} + U_ix_t) \\
  f_t =    & \sigma(W_fh_{t-1} + U_fx_t)\\
  o_t = &\sigma(W_oh_{t-1} + U_ox_t)\\
  c_t = &f_t \odot c_t-1 + i_t \odot tanh(W_ch_{t-1} + U_cx_t)\\
  h_t =& o_t \odot tanh(c_t)
\end{array}
\end{equation}
where $i_t$,$f_t$ and $o_t$ are input, forget and output gate respectively. $\sigma(\cdot)$ is a sigmoid function, and $W_i$, $W_f$, $W_o$, $U_i$, $U_f$, $U_o$ are parameters. Bi-LSTM is a combination of a forward lstm and a backward lstm. Naturally, it can capture both previous and future contextual semantics in this argument. Bi-LSTM generates two vectors at each step: $\overrightarrow{h_t}$ and $\overleftarrow{h_t}$ at step \textit{t}. We view their concatenation $h_t$=[$\overrightarrow{h_t}$; $\overleftarrow{h_t}$] as the \textit{t}-th positional representation of this argument.      

In the implicit discourse relation classification task, a pair of arguments is formally expressed as $(arg_1,arg_2)$, in which the former argument is represented as $arg_1 = (w_1^1,w_2^1,...,w_m^1)$ and the latter is $arg_2 = (w_1^2,w_2^2,...,w_n^2)$. Where \textit{m} and \textit{n} indicate lengths of the arguments respectively. $w_1^1,...,w_m^1,w_1^2,...,w_n^2$ are words initialized with pre-trained $d_e$-dimensional word embeddings. We feed $arg_1$ and $arg_2$ into Bi-LSTM:
\begin{eqnarray}
h_i^1 &=& {\rm \textbf{Bi-LSTM}}(arg_1,i)\\
h_j^2 &=& {\rm \textbf{Bi-LSTM}}(arg_2,j)
\end{eqnarray}
where \textit{i} and \textit{j} indicate the \textit{i}-th position in $arg_1$ and the \textit{j}-th position in $arg_2$, so that $h_i^1$ and $h_j^2$ stand for the corresponding positional argument representation which has incorporated inner semantics.

\begin{figure*}[t]
\centering
\includegraphics[scale=0.4]{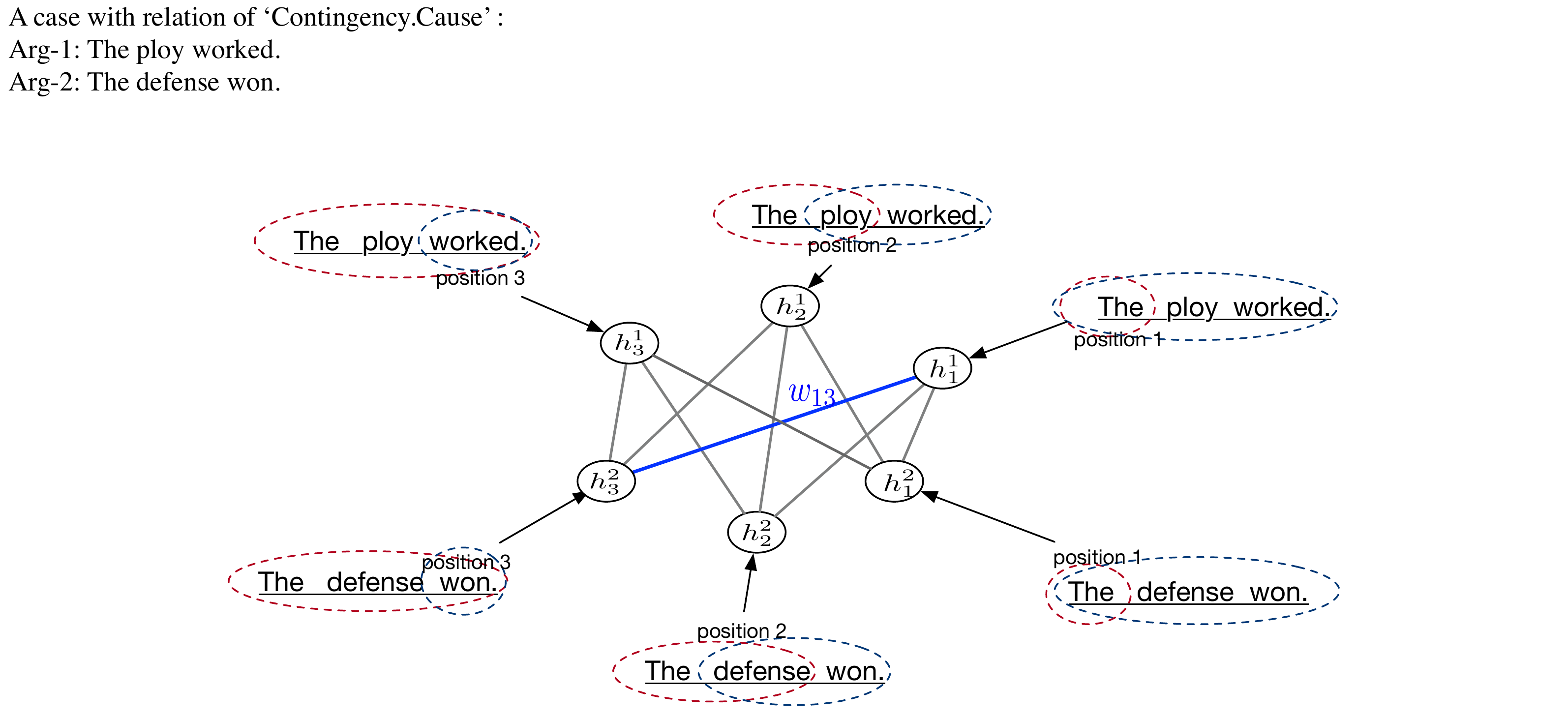}
\centering
\caption{The interaction graph. Here, white circles are the nodes initialized by the positional argument representations ($h_1^1$,...$h_3^2$). We only highlight one edge with weight $w_{13}$ for conciseness.
The red dotted lines enclose the semantics that the forward lstm capture at that position, while the purple dotted lines enclose semantics captured by the backward.}
\label{graph}
\end{figure*}

\subsection{Semantic Interaction Graph}
Representations of the arguments have been updated to $arg_1={({h_1^1},{ h_2^1},...,{h_m^1})}$ and $arg_2={({h_1^2}, {h_2^2},...,{h_n^2})}$. To represent inter-argument semantics structurally, we construct a semantic interaction graph \textit{G}=(\textit{V}, \textit{E}) based on these positional representations generated by Bi-LSTM, where \textit{V} and \textit{E} represent the sets of nodes and edges respectively. 

In order to show the meaning of the semantic interaction graph clearly, we pick a quite short case to give an example in Figure~\ref{graph}. Nodes in the graph correspond to the positions in $arg_1$ and $arg_2$. Since $arg_1$ has \textit{m} words and $arg_2$ has \textit{n}, we get $|V|=m+n$. Only the edges between positions from different arguments are set, so $|E|=m\times n$. There are no edges between positions from the same arguments, for this semantic interaction graph is mainly to model the inter-arguments semantic structure. 

The weight of the edge represents the degree of semantic relevance. The larger the weight, the stronger the relevance. Intuitively, we utilized a similarity function to compute the semantic relevance between the nodes from different arguments. Among kinds of similarity functions, we choose cosine which is common but effective. So, given a $arg_1$'s node $h_i^1$ and a $arg_2$'s node $h_j^2$, the weight $w_{ij}$ of the edge is calculated as follows:
\begin{equation}
{w_{ij}} = \frac{{h_i^1}^T h_j^2} {{\left\| {h_i^1} \right\|}\cdot {\left\| {h_j^2} \right\|}}
\label{squash}
\end{equation}
where ${\left\| \cdot \right\|}$ stands for the L2 norm. 
\citeauthor{Chen2016ImplicitDR}~\shortcite{Chen2016ImplicitDR} uses a similar but gated method to calculate the weights and obtain a score matrix, then feed it into classifier directly after pooling. On the contrary, our method extracts the deeper interactive features from the graph structure as described in the next section. 

\subsection{Graph Convolutional Layer}
After building the semantic interaction graph, we extract deeper interactive representations through a one-layer GCN. GCN is a multi-layer neural network that can encode the graph structure. It induces new representations of nodes by absorbing information of their neighbor nodes along the weighted edges. One layer GCN encodes only information about immediate neighbors. In this way, the nodes from one argument automatically pick out the semantic information of the nodes from the other argument, conducting a concise interactive process. We next introduce the details of graph convolution. 

Given the graph \textit{G}=(\textit{V}, \textit{E}), whose adjacent matrix is $A \in\mathbb{R}^{|V|\times|V|}$, we add self-connections on it so that the adjacent matrix tends to be $\tilde{A} = A+{I_N}$, where $I_N$ is the identity matrix. We define the order of nodes in adjacent matrix as the nodes from $arg_1$ are before the nodes from $arg_2$, then $\tilde{A}$ can be written as follows:
\begin{equation}      
 \tilde{A}=\left[         
  \begin{array}{cc}   
    I^{m\times m}& M\\  
     M^T& I^{n\times n }\\
  \end{array}
\right]                
\end{equation}
where,
\begin{equation}      
M=\left[         
  \begin{array}{cccc}   
    w_{11} &w_{12}  &... &w_{1n} \\  
    w_{21}& w_{22} &... &w_{2n} \\
     ...&...  &... &...  \\
    w_{m1} &w_{m2} &...&w_{mn}\\  
  \end{array}
\right]                
\end{equation}
Here, $I^{m\times m}$ and $I^{n\times n}$ are the identity matrix. $w_{ij}$ stands for the weight of the edge between \textit{i}-th position of $arg_1$ and the \textit{j}-th position of $arg_2$. The degree matrix is represented as $\tilde{D}$, where $\tilde{D}_{ii} = \sum_j\tilde{A}_{ij}$. Simultaneously, we concatenate the embedding of each node to a matrix $X={[h_1^1,...,h_m^1,...h_1^2,...,h_n^2]}^T\in\mathbb{R}^{|V|\times d_h}$, where
$d_h$ is twice the dimension of LSTM's hidden state. Then we obtain the new node feature matrix $X^\mathbb{g}$ through a layer of convolution, which has incorporated the structured interactive information.
\begin{equation}
X^\mathbb{g} = \sigma (\tilde{D}^{-\frac{1}2}\tilde{A}\tilde{D}^{-\frac{1}2}XW^\mathbb{g})
\end{equation}
Here, $\sigma(\cdot)$ is an activation function, such as the $RELU\mathbb{(\cdot)} = max(0,\cdot)$. $W^\mathbb{g}\in\mathbb{R}^{|V|\times d_g}$ is the weight matrix, so we obtain $X^\mathbb{g}\in\mathbb{R}^{|V|\times d_g}$ in which $d_g$ stands for the dimension of this convolutional layer. 

\subsection{Classification Layer}
As the last step of our framework, we infer a discourse relation type on the basis of $X^\mathbb{g}$ that has incorporated graph information. 

\paragraph{\bf Concat pooling:} Since the relation between two arguments is determined by some strong semantic signals in $X^\mathbb{g}$, we use max-pooling and mean-pooling to remove the redundant information. Given $X^\mathbb{g}=[(g_1^1),...,(g_m^1),...(g_1^2),...,(g_n^2)]^T$:
\begin{equation}
X^{\mathbb{c}} = [maxpool(X^\mathbb{g}),meanpool(X^\mathbb{g})]
\end{equation}
here, $X^{\mathbb{c}}\in \mathbb{R}^{1\times 2d_g}$, and $[ \cdot ]$ represents concatenating.

\paragraph{\bf MultiLayer Perception:} Finally, $X^{\mathbb{c}}$ obtained by the pooling layer is fed into a two-layer MLP for relation classification. There is one full connection hidden layer with RELU activation and then a softmax output layer in the MLP classifier. The entire model is trained end-to-end, through minimizing the cross-entropy loss.

\section{Experiments}

\subsection{IDR datasets}
We evaluate our method on two datasets: the PDTB2.0 and the Chinese Discourse Treebank (CDTB)~\cite{Zhou2012PDTBstyleDA}, which are currently the largest corpora annotated with discourse structures in English and Chinese respectively.  

\paragraph{\bf PDTB2.0:} The PDTB2.0 dataset contains 16,224 implicit relation instances totally. The sense labels of discourse relations in PDTB are hierarchical with three levels, i.e., class, type, and sub-type. There are four classes on the top level: COMPARISON (Comp.), CONTINGENCY (Cont.), EXPANSION (Exp.) and TEMPORAL (Temp.). 
We adopt two experimental settings on the top-level discourse relation prediction for the evaluation: the one is “one-versus-all” binary classification, the other is multi-class classification. We make the same data splitting as the predecessors~\cite{Pitler2009AutomaticSP} do: sections 2-20 for training, sections 21-22 for testing and sections 0-1 for the validation set. Table 2 shows the distribution statistics of top-level relations. 

\noindent\paragraph{\bf CDTB:} The CDTB is annotated as the PDTB-style but consists of 10 different fine-grained relations.
We evaluate our model on the official train, development, test and blind-test sets of the CDTB provided to the CoNLL-2016 Shared Task participants. Following the predecessors~\cite{Rutherford2016RobustNN}, we use Accuracy as the evaluation metric, and treat ENTREL (entity relation) as implicit and exclude ALTLEX relation, so we perform a 9-way classification on the remaining nine relations. The Train/Dev/Test/Blind-Test set contains 7828/301/352/1486 instances, respectively.


\subsection{Experimental Setup}

We introduce the experimental setup including baselines and implementation details in this section. The baselines on PDTB are carefully divided into two categories for a comprehensive comparison: 

\begin{table} [t]
\centering
\scalebox{1.0}{
\begin{tabular}{p{2cm}<{\centering}p{1.3cm}<{\centering}p{1.3cm}<{\centering}p{1.3cm}<{\centering}}  
\hline
\textbf{Relation} & \textbf{Train} & \textbf{Dev}& \textbf{Test}\\ 
\hline\hline
\textbf{Comp.} & 1942 & 197 & 152\\ 
\textbf{Cont.} & 3342 & 295 & 279\\ 
\textbf{Exp.} & 7004  & 671&574\\ 
\textbf{Temp.}& 760 & 64 & 85\\
\hline
\end{tabular} 
}
\caption{The distribution statistics of four top-level classes in PDTB 2.0} 
\end{table}

\begin{table*}[t]  
\centering
\scalebox{1.0}{
\begin{tabular}{c|cp{1.5cm}<{\centering}p{1.5cm}<{\centering}p{1.5cm}<{\centering}p{1.5cm}<{\centering}}  
\hline
\textbf{External Resource} &{\textbf{\multirow{1}*{Model}}} & {\textbf{\multirow{1}*{Comp.}}} & {\textbf{\multirow{1}*{Cont.}}}& {\textbf{\multirow{1}*{Exp.}}}& {\textbf{\multirow{1}*{Temp.}}}\\

\hline\hline
\textbf{\multirow{4}*{With}}
&\citeauthor{Qin2017AdversarialCN}~\shortcite{Qin2017AdversarialCN} & 40.87& 54.56 & 72.38 & 36.20 \\  
&\citeauthor{Lan2017MultitaskAN}~\shortcite{Lan2017MultitaskAN} & 40.73&  \textbf{58.96} &  72.47& 38.50 \\
&\citeauthor{dai2018improving}~\shortcite{dai2018improving} & 42.68 &55.17 & 68.94& \textbf{41.03}  \\
&\citeauthor{lei2018linguistic}~\shortcite{lei2018linguistic} & \textbf{43.24}&  57.82&\textbf{72.88}& 29.10 \\
\hline\hline
\textbf{\multirow{5}*{Without}}
&Bi-LSTM &37.14  &52.96  & 69.72 & 34.78 \\ 
&Bi-LSTM+Attention &39.25 & 54.21 &70.48  & 37.06  \\
&\citeauthor{Chen2016ImplicitDR}~\shortcite{Chen2016ImplicitDR}&40.17&54.76&-&31.32\\
&\citeauthor{Lan2017MultitaskAN}~\shortcite{Lan2017MultitaskAN} & 38.15&  \textbf{56.07}&70.53& 36.72 \\

\cline{2-6}
&\textbf{SGCN$\left(ours\right)$} & \textbf{43.90}& 55.23 & \textbf{71.83} & \textbf{43.80} \\  

\hline
\end{tabular} 
}
\caption{ Comparisons of F1 scores (\%) for the multiple binary classification on the top level classes in PDTB 2.0 
}
\end{table*}

\paragraph{\bf Models with external resources:} such as the external corpus, the additional annotations, and the parser.
\begin{itemize}
\item \citeauthor{Qin2017AdversarialCN}~\shortcite{Qin2017AdversarialCN}: A novel adversarial model that enables an adaptive imitation scheme through competition between the implicit network and a rival feature discriminator, incorporating implicit connectives which are the additional annotation in PDTB 2.0.
\item \citeauthor{Lan2017MultitaskAN}~\shortcite{Lan2017MultitaskAN}: A novel multi-task attention-based model, where the proposed multi-task framework can learn external knowledge from corpora of similar tasks for better classification. 
\item \citeauthor{dai2018improving}~\shortcite{dai2018improving}: This method models the overall paragraph-level discourse structure, using the inter-dependencies between the discourse units to predict a sequence of discourse relations in a paragraph. Their work utilizes both implicit and explicit corpora.
\item \citeauthor{lei2018linguistic}~\shortcite{lei2018linguistic}: This method designs linguistic features to enhance classification, which requires lots of feature engineering work and uses many external resources such as Stanford Parser, the MPQA Corpus\footnote{https://mpqa.cs.pitt.edu/} and so on.
\end{itemize}

\begin{table} [t]
\centering
\scalebox{1.0}{
\begin{tabular}{c|p{3.5cm}<{\centering}p{1cm}<{\centering}p{1cm}<{\centering}}  
\hline
\textbf{External}&\multirow{2}*{\textbf{Model}} & \multirow{2}*{\textbf{F1}} & \multirow{2}*{\textbf{Acc}}\\ 
\textbf{Resource}&&  &\\ 
\hline
\hline
\textbf{\multirow{4}*{With}}
&\citeauthor{Qin2017AdversarialCN}~\shortcite{Qin2017AdversarialCN} &-&- \\  
&\citeauthor{Lan2017MultitaskAN}~\shortcite{Lan2017MultitaskAN} & \textbf{47.80}& \textbf{57.39}  \\
&\citeauthor{dai2018improving}~\shortcite{dai2018improving} &47.56&56.88 \\
&\citeauthor{lei2018linguistic}~\shortcite{lei2018linguistic} & 47.15& - \\
\hline
\hline
\textbf{\multirow{4}*{Without}}
&Bi-LSTM &36.79& 54.09  \\ 
&Bi-LSTM+Attention &42.24&55.62\\
&\citeauthor{Chen2016ImplicitDR}~\shortcite{Chen2016ImplicitDR}&-&-\\
&\citeauthor{Lan2017MultitaskAN}~\shortcite{Lan2017MultitaskAN} &45.57& \textbf{57.55}  \\

\cline{2-4}
&\textbf{SGCN$\left(ours\right)$} & \textbf{47.38}& 56.79\\  

\hline
\end{tabular} 
}
\caption{Comparisons of F1 scores (\%) and accuracy (\%) for 4-class classification on the top level classes in PDTB 2.0 } 
\end{table}

\begin{table} [t]
\centering
\scalebox{1.0}{
{
\begin{tabular}{p{4.0cm}<{\centering}p{1.5cm}<{\centering}p{1.5cm}<{\centering}}  
\hline
\textbf{Model} & \textbf{Test}& \textbf{Blind} \\ 
\hline
\hline
Bi-LSTM &71.94&62.88\\ 
Bi-LSTM+Attention &72.63&64.07\\
\citeauthor{Rutherford2016RobustNN}~\shortcite{Rutherford2016RobustNN}&70.47&63.38\\
\citeauthor{Wang2016TwoES}~\shortcite{Wang2016TwoES}&72.42&60.52\\
\citeauthor{Rnnqvist2017ARN}~\shortcite{Rnnqvist2017ARN}&73.01&-\\
\textbf{SGCN}(\textit{ours}) &\textbf{73.85}&\textbf{66.02}\\
\hline
\end{tabular} 
}
}
\caption{The accuracy of 9-class classification on CDTB} 
\end{table}

\noindent\paragraph{\bf Models without external resources:}
\begin{itemize}
\item Bi-LSTM: We use a one-layer Bi-LSTM to encode the pair of arguments, then concatenate their last hidden state and feed it to a two-layer MLP for classification. 
\item Bi-LSTM+Attention: We apply attention mechanism~\cite{Cho2014LearningPR} on Bi-LSTM to model the interaction between arguments. After getting $arg_2$-aware representation of $arg_1$ and $arg_1$-aware representation of $arg_2$, we feed their concatenation into the MLP. 
\item \citeauthor{Chen2016ImplicitDR}~\shortcite{Chen2016ImplicitDR}: This model adopts a gated relevance network to capture the interactive information between word pairs, and then aggregate them using a pooling layer to select the most informative interactions.
\end{itemize}

Among the models mentioned above, some are exploiting the role of semantic interaction, such as Bi-LSTM+Attention, \citeauthor{Chen2016ImplicitDR}~\shortcite{Chen2016ImplicitDR}, and \citeauthor{Lan2017MultitaskAN}~\shortcite{Lan2017MultitaskAN}, and others are novel methods that have recently achieved the best results. 





\noindent\paragraph{\bf Implementation details:} The English word embeddings are initialized with 300-dimensional-GloVe vectors~\cite{Pennington2014GloveGV}, and the Chinese word embeddings are initialized with the Financial vectors\footnote{https://github.com/Embedding/Chinese-Word-Vectors}. The hyper-parameters have been optimized over the development set. We set the size of LSTMs'hidden states to 128, the GCN layer to 100, two-layer MLP's first hidden layer to 64 while its output layer is determined by the number of relations. The minibatch size is set to 64. We train our model using Adam~\cite{Kingma2014AdamAM} with gradient clipping range $[-5.,5.]$. We set the learning rate to $1e^{-2}$, and decayed after every epoch by a factor of $0.9$.  

\subsection{Main Results}
\paragraph{\bf Results on PDTB 2.0:} Table 3 shows the evaluation results for One-versus-all of the top level classes on the benchmark of PDTB 2.0. 
Compared with those methods that have used external resources~\cite{Qin2017AdversarialCN,Lan2017MultitaskAN,dai2018improving,lei2018linguistic}, our model still achieves F1 improvements of 1.53\% on Comp. and 7.2\% on Temp., the numbers of samples belonging to which are the two least in all classes as shown in Table 2. These results demonstrate the effectiveness of our method, especially in deficient data scenarios. On the other two classes, Exp. and Cont., we also achieve comparable performance.
The proposed SGCN shows obvious advantages in most relations, when compared with the methods that do not use external resources, including the Bi-LSTM+Attention and ~\citeauthor{Chen2016ImplicitDR}~\shortcite{Chen2016ImplicitDR}, which also exploit inter-argument semantics. 
Different from these approaches, our approach models the semantic interactions structurally and extract the deeper interactive features from the learned semantic structure. This is the main reason why our method outperforms them in most of the relations. We also report the results of the 4-class classification for more comparisons with prior works. Table 4 reports the macro-averaged F1 scores and Accuracy. We can see that our method achieve the highest F1 score compared with methods which don't use external resource and obtain comparable scores with the best-performing methods using external resources. 

\noindent\paragraph{Results on CDTB:} Following \citeauthor{Rutherford2016RobustNN}~\shortcite{Rutherford2016RobustNN}, we use Accuracy as the evaluation metric on the benchmark of CDTB. As Table 5 shows, our model outperforms all of the previous systems and achieves Accuracy improvements of 1.2\% on the Test set  and 4.2\% on the Blind-Test set. Performances on these two test sets suggest the robustness of our approach and its ability to generalize to unseen data.

\subsection{Effects of GCN setting}
In this section, we investigate the effects of different GCN settings. Table 6 shows the performance of top-level One-versus-all Classifications on the test set of PDTB 2.0 in different GCN settings. 

\begin{table} [t]
\centering
\scalebox{1.0}{
{
\begin{tabular}{ccccc}  
\hline
\textbf{Size} & \textbf{Comp.} &  \textbf{Cont.} & \textbf{Exp.} &  \textbf{Temp.}\\ 
\hline\hline
50d &41.78&51.89&71.40&41.14 \\ 
100d &\textbf{43.90}&\textbf{55.23}&\textbf{71.83}&\textbf{43.80}\\
150d &42.81&54.32&71.80&43.21\\
200d &40.39&52.73&71.22&39.02\\
\hline
\end{tabular} 
}
}
\caption{Results of four binary classification with different GCN sizes. `d' stands for the dimension of convolutional layer.} 
\end{table}
\paragraph{\bf GCN-size:} Different sizes are set up for the graph convolutional layer. We can see that the best performance is achieved when we use a moderate size `100d' of graph convolutional layer. We conjecture that a moderately sized GCN can avoid losing too much information, as well as being immune to redundant information. 

\subsection{Case Study}
In this section, we adopt a visualization way to further validate the ability of the proposed SGCN to capture effective interactive semantics. 
We define an interaction score matrix for visualization. Suppose there are $m$ nodes from $arg_1$ and $n$ nodes from $arg_2$. For every two nodes from different arguments, we calculate the semantic interactive score of their representations through similarity function; then we obtain a matrix $M\in\mathbb{R}^{m\times n}$ as Eq.(5).  
In order to show what information in the argument pairs is captured in SGCN clearly, we then calculate $M^{'}\in \mathbb{R}^{m\times n}$ in the way we obtain $M$. Which is different, the new node representations after the graph convolution which carries deeper interactive information are used. We pick a real case in PDTB2.0 as an example:\\

\textit{\textbf{arg-1: There was a general feeling that we'd seen the worst.}}

\textit{\textbf{arg-2: The resignation came as a great surprise.}}\\

\begin{figure}[t]
\centering
\includegraphics[scale=0.35]{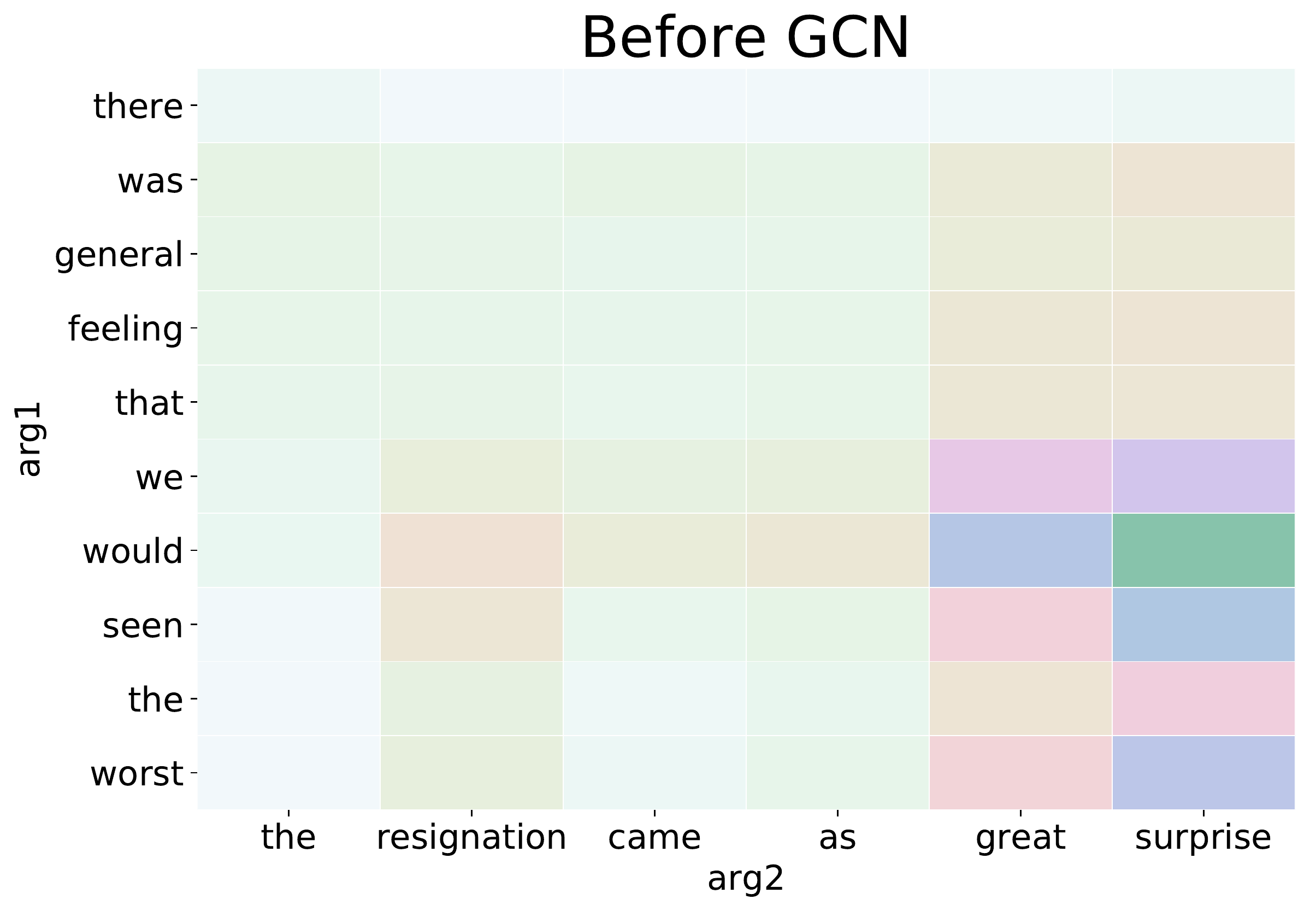}
\centering
\caption{The interactive score matrix calculated by the shallow interactive representations before the GCN. The darker the color, the bigger the higher the interactive score. Here, ``great",`` surprise" get two highest interactive scores with arg-1.} 
\label{shallow}
\end{figure}

The relation type of this case is \textit{Contingency} and the implicit connective annotated by human is ``so", while the word pairs (\textit{worst}, \textit{great}) and (\textit{worst}, \textit{surprise}) may wrongly trigger a \textit{Comparison} relation if we fail to capture the deeper interaction of whole arguments. For this case, we visualize two matrix $M$ and $M'$ to show what information is captured before and after the GCN, respectively.

Figure~\ref{shallow} shows the visualization of the matrix $M$, which is computed based on the node representations before the GCN. We can see that the positions of ``\textit{great}" and ``\textit{surprise}" in arg-2 get two highest interactive scores with arg-1, while the score of ``\textit{resignation}" is much lower than them. It shows that the shallow interactive features before GCN fail to get the point of semantics. High score of (\textit{worst}, \textit{great}) will lead to wrong inference.

Figure~\ref{deep}  shows the visualization of the matrix $M'$, which is computed based on the new node representations after the GCN. As shown in Figure ~\ref{deep}, the position of ``\textit{resignation}" obtains a significantly higher score than others. In the case that the overall scores of ``\textit{great}" and ``\textit{surprise}" are not very low, the scores of (\textit{worst}, \textit{great}) and (\textit{worst}, \textit{surprise}) are almost the lowest in the whole matrix, which avoids misleading to a \textit{Comparison} relation. We can see that the deeper interactive features obtained through graph convolutional layer are more effective for the classification.

\begin{figure}[t]
\centering
\includegraphics[scale=0.35]{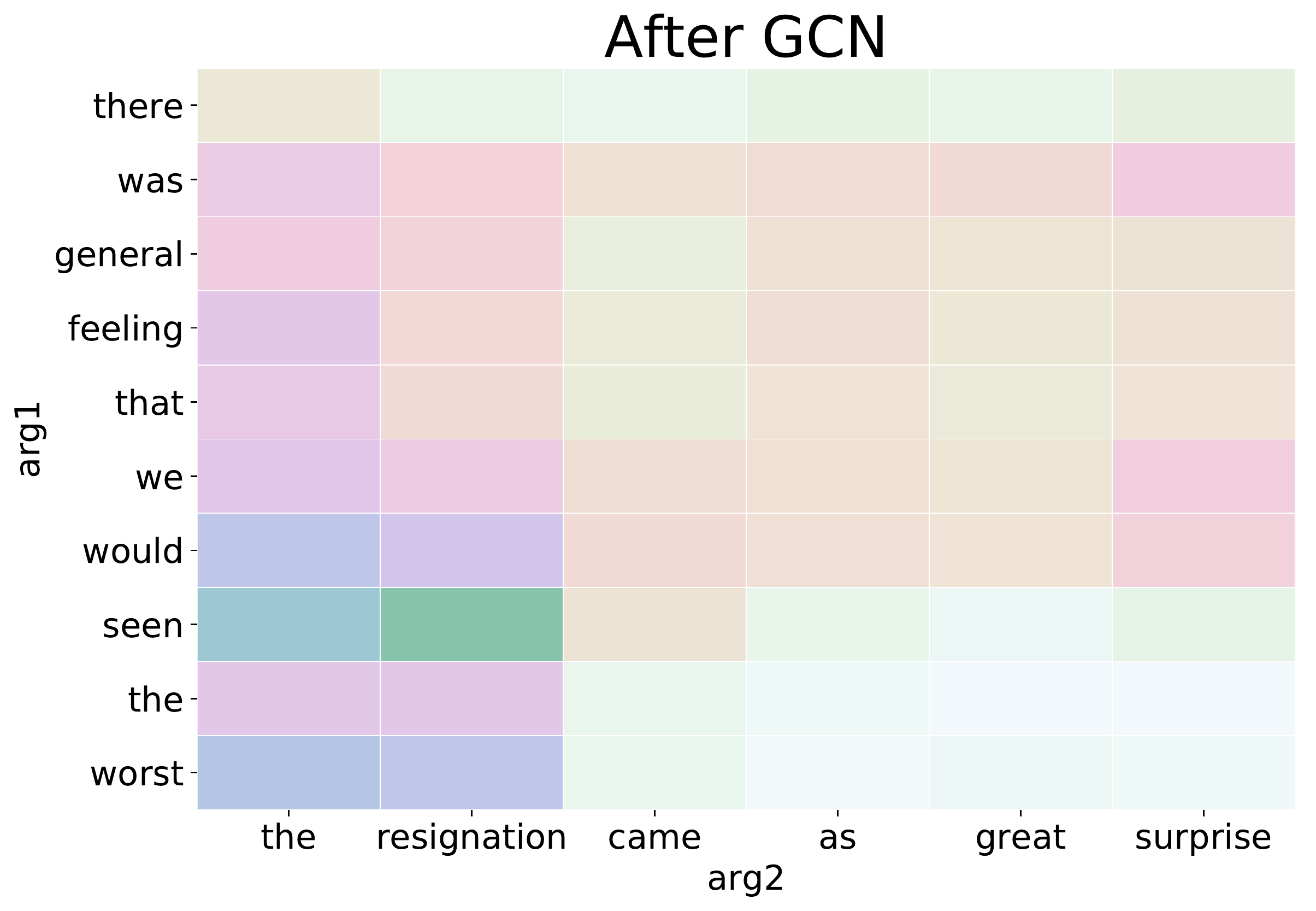}
\centering
\caption{The interactive score matrix calculated by the deeper interactive representations after the GCN. The darker the color, the bigger the higher the interactive score. Here, ``the  resignation" gets much higher interactive scores than ``great surprise".} 

\label{deep}
\end{figure}

\section{Conclusion}
We propose a novel and effective Semantic Graph Convolutional Network (SGCN) for implicit discourse relation classification, which can model the inter-arguments semantics structurally and capture the deeper interactive semantics. The experiments on PDTB 2.0 and CDTB demonstrate the superiority of the proposed model to previous state-of-the-art systems. We also compute the interactive score matrix before and after the GCN respectively for visualization, which demonstrates that the SGCN successfully captures useful interactive semantics. To the best of our knowledge, this is the first work that exploits a graphical structure to model the semantic interaction between arguments. 

For future work, we would like to exploit external resources to further boost the performance of our SGCN for implicit discourse relation classification, such as an entity-augmented graph. 
\newpage

\bibliographystyle{aaai}
\bibliography{main}

\else
\begin{center}
\textbf{Warning:} The DVI version of this paper may be corrupted.  If possible, use the PDF version.
\end{center}
\fi

\end{document}